\documentclass[preprint,12pt]{elsarticle}

\usepackage{amssymb}
\usepackage{lineno,hyperref}
\usepackage{multirow}
\usepackage{color}
\graphicspath{{figures/}}
\journal{ }

\begin{document}
\begin{frontmatter}

\title{Infrared and Visible Image Fusion with ResNet and zero-phase component analysis}

\author[a]{Hui Li}

\author[a]{Xiao-jun Wu\corref{correspondingauthor}}

\author[b]{Tariq S. Durrani}

\address[a]{Jiangsu Provincial Engineering Laboratory of Pattern Recognition and Computational Intelligence,Jiangnan University, 214122 Wuxi, China \fnref{address1}}
\address[b]{Department of Electronic and Electrical Engineering, University of Strathclyde, Glasgow G1 1XW, UK \fnref{address2}}

\cortext[correspondingauthor]{Corresponding author email: wu\_xiaojun@jiangnan.edu.cn}

\begin{abstract}
Feature extraction and processing tasks play a key role in Image Fusion, and the fusion performance is directly affected by the different features and processing methods undertaken. By contrast, most of deep learning-based methods use deep features directly without feature extraction or processing. This leads to the fusion performance degradation in some cases. To solve these drawbacks, we propose a deep features and zero-phase component analysis (ZCA) based novel fusion framework is this paper. Firstly, the residual network (ResNet) is used to extract deep features from source images. Then ZCA is utilized to normalize the deep features and obtain initial weight maps. The final weight maps are obtained by employing a soft-max operation in association with the initial weight maps. Finally, the fused image is reconstructed using a weighted-averaging strategy. Compared with the existing fusion methods, experimental results demonstrate that the proposed framework achieves better performance in both objective assessment and visual quality. The code of our fusion algorithm is available at \url{https://github.com/hli1221/imagefusion_resnet50}.

\end{abstract}


\end{frontmatter}

%\modulolinenumbers[5]
%\pagewiselinenumbers
%\linenumbers

%% main text
\section{Introduction}
\label{introduction}

Infrared and visible image fusion is a frequently occuring requirement in image fusion, and the fusion methods for this work are widely used in many applications. These algorithms combine the salient features of source images into a single image\cite{1}.

For decades, signal processing algorithms\cite{2} \cite{3} \cite{4} \cite{5} were the most propular feature extraction tools in image fusion tasks. In 2016, a fusion method based on two-scale decomposition and saliency detection was proposed by Bavirisetti et al.\cite{6}. The base layers and detail layers were extracted by a mean filter and a median filter. The visual salient features were used to obtain weight maps. The fused image was then reconstructed by combining these three parts.

In recent years, representation learning-based fusion methods have attracted great attention and exhibited state-of-the-art fusion performance. In the sparse representation(SR) domain, Zong et al.\cite{7} proposed a novel medical image fusion method based on SR. In their paper, the sub-dictionaries are learned by Histogram of Oriented Gradients (HOG) features. Then the fused image is reconstructed by $l_1$-norm and the max selection strategy. In addition, the joint sparse representation\cite{8}, cosparse representation\cite{9}, pulse coupled neural network(PCNN)\cite{10} and shearlet transform\cite{11} are also applied to image fusion, which incorporate the SR. 

In other representation learning domains, for the first time, the low-rank representation(LRR) was applied into image fusion tasks by Li et al.\cite{12}. In \cite{12}, they use HOG and dictionary learning method to obtain a global dictionary. The dictionary is then used in LRR and the fused low-rank coefficients are obtained by using an $l_1$-norm and choose-max strategy. Finally, the fused image is reconstructed using the global dictionary and LRR. 

%For infrared and visible image fusion, Li et.al\cite{13} also proposed an effective and simple algorithm based on latent low-rank representation(LatLRR). Here the source images are decomposed into low-frequency and high-frequency coefficients by LatLRR and the fused image is reconstructed by using a weighted-averaging strategy. 

Although these representation learning-based methods exhibit good fusion performance, they still have two main drawbacks: 1) It is difficult to learn an effective dictionary offline for representation learning-based methods; 2) The time efficiency of representation learning-based methods is very low, especially, when the online dictionary learning strategies are used in fusion algorithms. So recently, the fusion algorithms have been improved in two aspects: time efficiency and fusion performance.

In the last two years, deep learning has been applied to image fusion tasks shown to achieve better fusion performance and time efficiency than non-deep learning based methods. Most of the deep learning-based fusion methods just treat deep learning as feature extraction operation and use deep features which are obtained by a fixed network to reconstruct the fused image. In \cite{14}, a convolutional sparse representation(CSR) based fusion method was proposed by Yu Liu et al. The CSR is used to extract features which are obtained by different dictionaries. In addition, Yu Liu et al.\cite{15} also proposed an algorithm based on convolutional neural network(CNN) for multi-focus images. Image patches which contain different blur versions of input image are utilized to train the network. And a decision map is obtained by this network. Finally, the decision map and the source images are used to reconstruct the fused image. The obvious drawback of these two methods is that they are just suitable for the multi-focus image fusion task.

In ICCV 2017, Prabhakar et al.\cite{16} proposed a simple and efficient method based on CNN for the exposure fusion problem(DeepFuse). In their method, a siamese network architecture where the weights are tied is used to construct the encoding network. Then two feature map sequences are obtained by encoding. And they are fused by an addition strategy. The final fused image is reconstructed by a decoding network which contains three CNN layers. This network is not only suitable for the exposure fusion problem, it also achieves good performance in other fusion tasks. However, the architecture is too simple and the information contained in deep network may not have been fully utilized.

As a follow-on, Li et al.\cite{41} proposed a novel fusion network based on DeepFuse and densenet\cite{42} which is named DenseFuse. This fusion network contains three parts: encoder, fusion layer and decoder. Thee encoder network is combined by convolutational layer and denseblock. And the fusion layer is also called fusion strategy in image fusion tasks.

In addition, Li et al.\cite{17} also proposed a VGG-based\cite{18} fusion method which uses a deep network and multi-layer deep features. Firstly, the base parts and the detail content are obtained by a decomposed method from source images. The base parts are fused by weighted-averaging strategy. And the fixed VGG-19 network, which is trained by ImageNet, is used to extract multi-layer deep features from detail content. Then weight maps are calculated by soft-max operator and multi-layer deep features. Several candidates fused detail content is obtained by weight maps. The choose-max is used to construct the final weight maps for the detail content. The final weight maps are utilized to obtain fused detail content. Finally, the fused image is reconstructed by combining the fused base part and the detailed content.

Although information from the middle layers is used by VGG-based fusion method\cite{17}], the multi-layer combining method is still too simple, and much useful information is lost in feature extraction. This phenomenon gets worse when the network is deeper.

To solve these problems, we propose a fusion method to fully utilize and process the deep features. In this article, a novel fusion framework based on residual network(ResNet)\cite{19} and zero-phase component analysis(ZCA)\cite{20} is proposed for infrared and visible image fusion task. The ResNet which is fixed is used to obtain the deep features from source images. Due to the architecture of ResNet, the deep features already contain multi-layer information, so we just use the output which is obtained by single layer. Then ZCA operation is utilized to project deep features into sparse domain and initial weight maps are obtained by $l_1$-norm operation. We use bicubic interpolation to reshape the initial weight maps to source image size. And the final weight maps are obtained by soft-max operation. Finally, the fused image is reconstructed by final weight maps and source images.

In Section\ref{related}, we review related work while Section\ref{proposed} we describe our fusion algorithm. The experimental results are shown in Section\ref{experiment}. Finally, Section\ref{con} draws the conclusions to the paper.

\section{Related work}
\label{related}

\subsection{Deep residual network (ResNet)}
\begin{figure}[ht]
\centering
\includegraphics[width=0.3\linewidth]{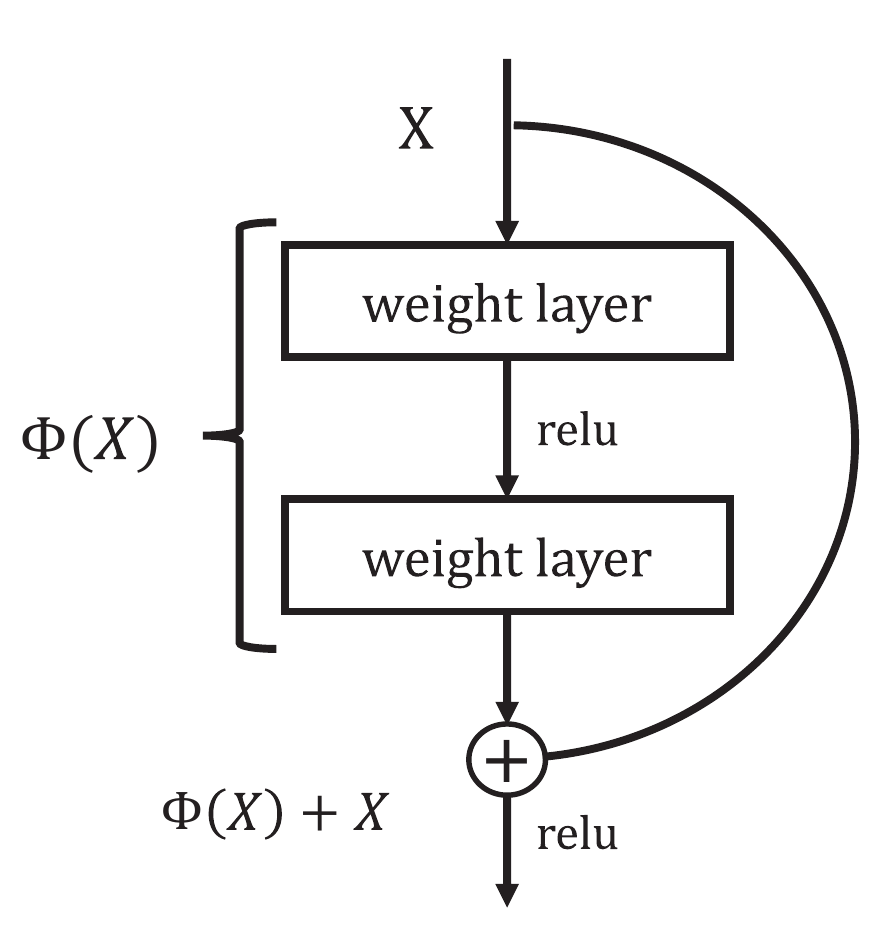}
\caption{The architecture of residual block.}
\label{fig:resblock}
\end{figure}

%\textbf{Deep residual network (ResNet)}. 
In CVPR 2016, He et al.\cite{19} proposed a novel network architecture to address the degradation problem. With the shortcut connections and residual representations, their nets were easier to optimize than previous networks and offered better accuracy by increasing the depth. The residual block architecture is shown in Fig.\ref{fig:resblock}.

$X$ indicates the input of net block, $\Phi{(X)}$ denotes the network operation which contains two weight layers, and ``relu'' represents the rectified linear unit. The output of residual block is calculated by $\Phi{(X)}+X$. With this structure, the multi-layer information is utilized. Furthermore, in image reconstruction tasks \cite{21} \cite{22} \cite{23}, the performance gets better by apply the residual block. We also use this architecture in our fusion methods.

\subsection{Zero-phase component analysis(ZCA)}
%\textbf{Zero-phase component analysis(ZCA)}.
In \cite{20}, Kessy et al. analyzed the whitening and decorrelation by ZCA operation. ZCA operation is used to project a random vector into a irrelevant sub-space which is also named whitening. In image processing field, ZCA is a useful tool to process the features, which can obtain useful features to improve algorithm performance. We will introduce ZCA operation briefly.

Let $X=(x_1,x_1,\cdots,x_d )^T$ indicate the $d$-dimensional random vector and $u=(u_1,u_1,\cdots,u_d )^T$ represent the mean values. And the covariance matrix $Co$ will be calculated by $Co=(X-u)\times(X-u)^T$. Then the Singular Value Decomposition(SVD) is utilized to decompose $Co$, as shown in Eq.\ref{eq:1},
\begin{eqnarray}\label{eq:1}
  	[U,\Sigma,V]=SVD(Co) \\
    s.t.,Co=U\Sigma V^T \nonumber
\end{eqnarray}

Finally, the new random vector $\hat{X}$ is calculated by Eq.\ref{eq:2},
\begin{eqnarray}\label{eq:2}
  	\hat{X}=U(\Sigma+\epsilon I)^{-\frac{1}{2}}U^T\times X
\end{eqnarray}
where $I$ denotes the identity matrix, and $\epsilon$ is a small value avoiding bad matrix inversion.

\subsection{ZCA utilization in image style transfer}
%\textbf{ZCA utilization in image style transfer}. 
Recently, ZCA is also utilized in image style transfer task which is one of the most popular in the image processing field.

Li et al.\cite{24} proposed a universal style transfer algorithm using ZCA operation to transfer the style of artistic image into content image. The encoder network is used to obtain the style features($f_s$) and content features($f_c$). Then authors use ZCA operation to project $f_s$ and $f_c$ into the same space. The final transferred features will be obtained by a coloring transform method which is a reverse operation to the ZCA operation. Finally, the styled image is obtained by transferred features and a decoder network.

In addition, in CVPR 2018, Lu et al.\cite{25} also use ZCA operation in their style transfer method. The VGG network is utilized to extract image features, and ZCA is used to project features into the same space. Then transferred features are obtained by a reassembling operation based on patches. Finally, the transferred features and a decoder network is trained by MSCOCO\cite{26} dataset and utilized to reconstruct the styled image.

From above style transfer methods, ZCA is a powerful tool to process image features, especially in the image reconstruction task. It projects image features into a sub-space, which makes features easy to classify and reconstruct. Inspired by these methods, we also apply the ZCA operation into image fusion task.

\section{The Proposed Fusion Method}
\label{proposed}

In this section, the proposed fusion method is introduced in detail.

\begin{figure}[ht]
\centering
\includegraphics[width=1\linewidth]{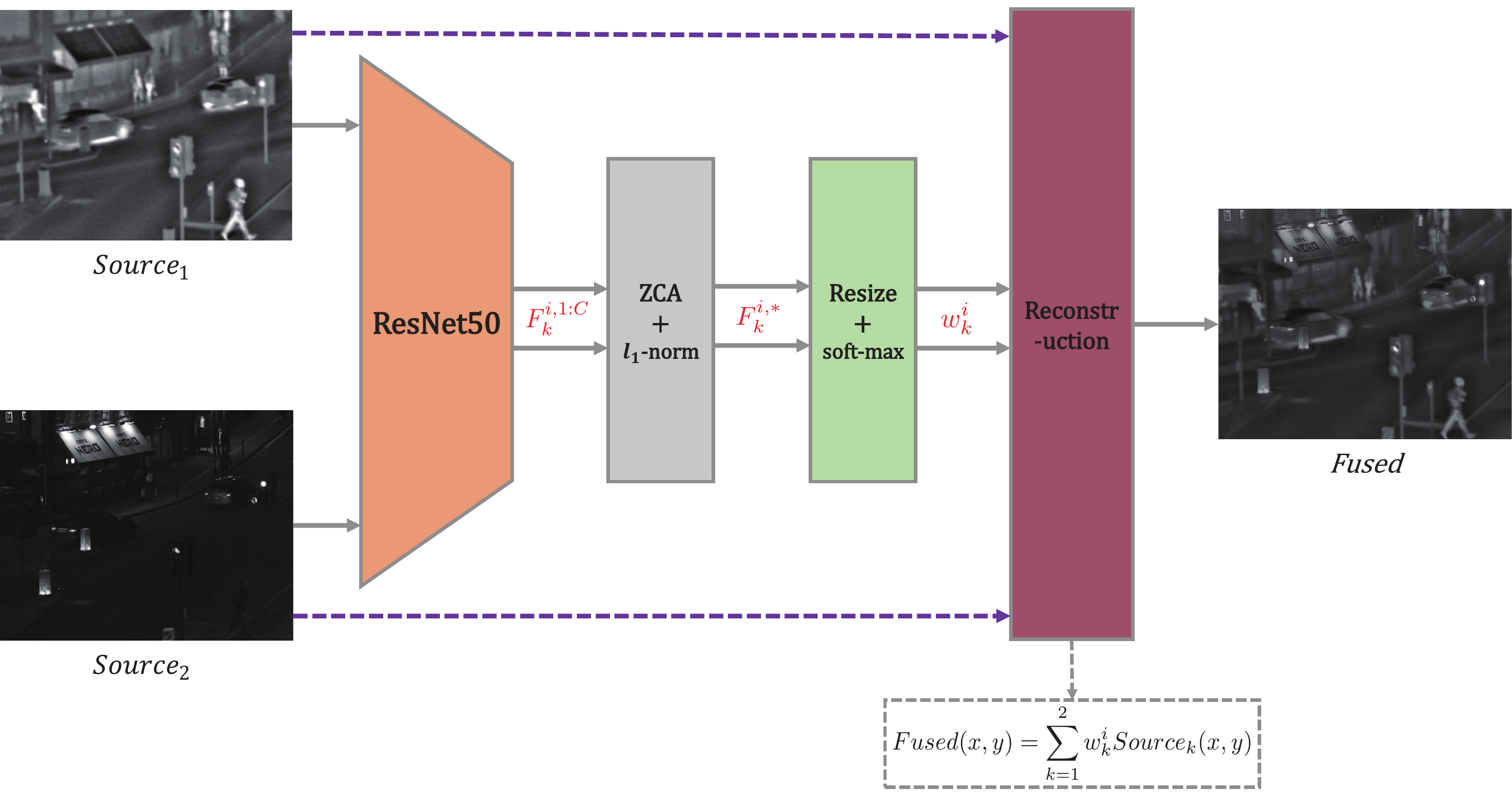}
\caption{The framework of proposed method.}
\label{fig:framework}
\end{figure}

Assuming there are $K$ preregistered source images, in our paper, the $K=2$. Note that the fusion strategy is the same for $K>2$. The source images are represented as $Source_k$,$k\in{\{1,2\}}$. The framework of the proposed fusion method is shown in Fig.\ref{fig:framework}.

In Fig.\ref{fig:framework}, the source images are indicated as $Source_1$ and $Source_2$, and ResNet50 contains 50 weight layers which include 5 convolutional blocks (conv1, conv2, conv3, conv4, conv5). The ResNet50 is a fixed network and trained by ImageNet\cite{27}, we use it to extract the deep features. And the output of $i$-th blocks are indicated by the deep features $F_k^{i,1:C}$ which contain C channels, $i\in{\{1,2,\cdots,5\}}$. We use ZCA and $l_1$-norm to process $F_k^{i,1:C}$ to obtained $F_k^{i,*}$. Then the weight maps $w_k^i$ are obtained by resize (bicubic interpolation) and soft-max operation. Finally, the fused image is reconstructed by use weighted-average strategy. In our paper we choose $i=4$ and $i=5$ to evaluate our fusion framework.

\subsection{ZCA operation for deep features}

\begin{figure}[ht]
\centering
\includegraphics[width=1\linewidth]{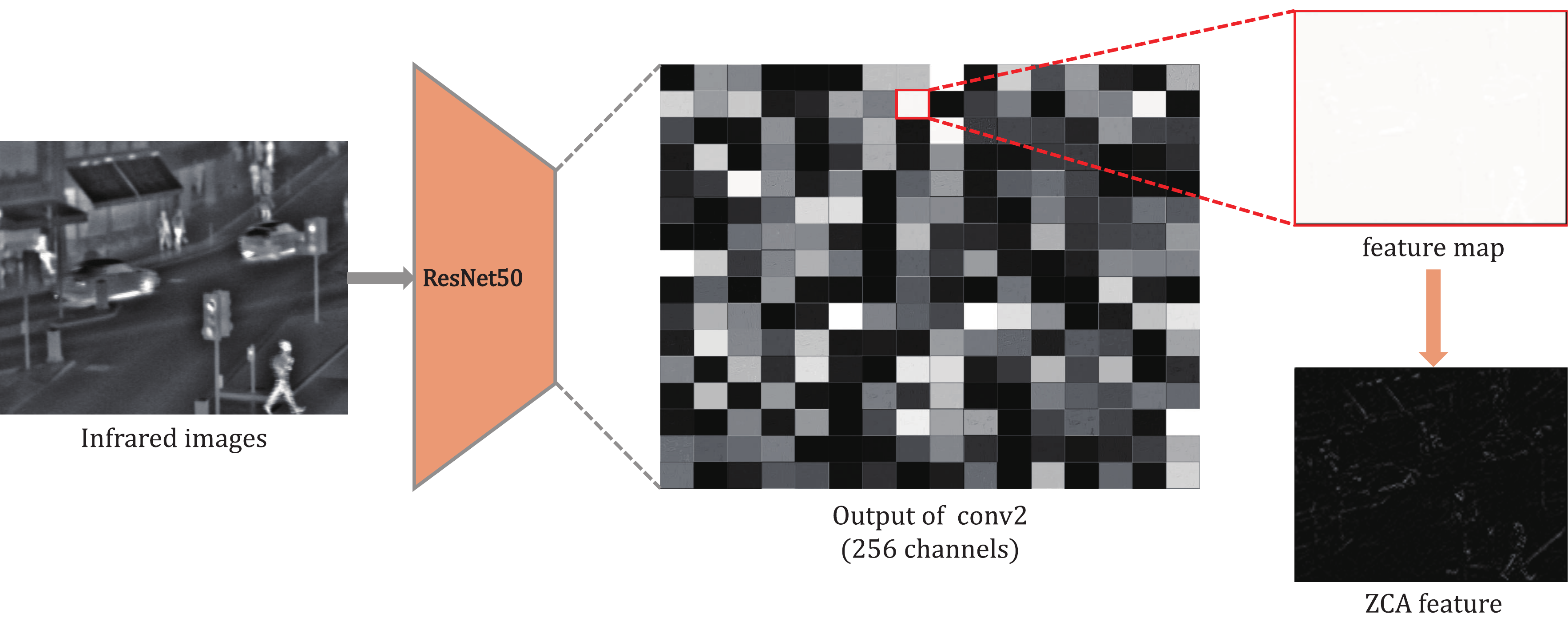}
\caption{ZCA operation for deep features.}
\label{fig:zca}
\end{figure}

As we discussed earlier, ZCA projects the original features into the same space, and the features become more useful for the next processing. The ZCA operation for deep features is shown in Fig.\ref{fig:zca}.

In Fig.\ref{fig:zca}, we choose the output of conv2 layer which contains 3 residual blocks as an example to introduce the influence of ZCA operation. Each block indicates one channel of the output. The original deep features have different orders of magnitude in each channel. We use ZCA to project original features into the same space. The features become more significant, as shown in Fig.\ref{fig:zca} (ZCA feature).

\subsection{ZCA and $l_1$-norm operations}

\begin{figure}[ht]
\centering
\includegraphics[width=1\linewidth]{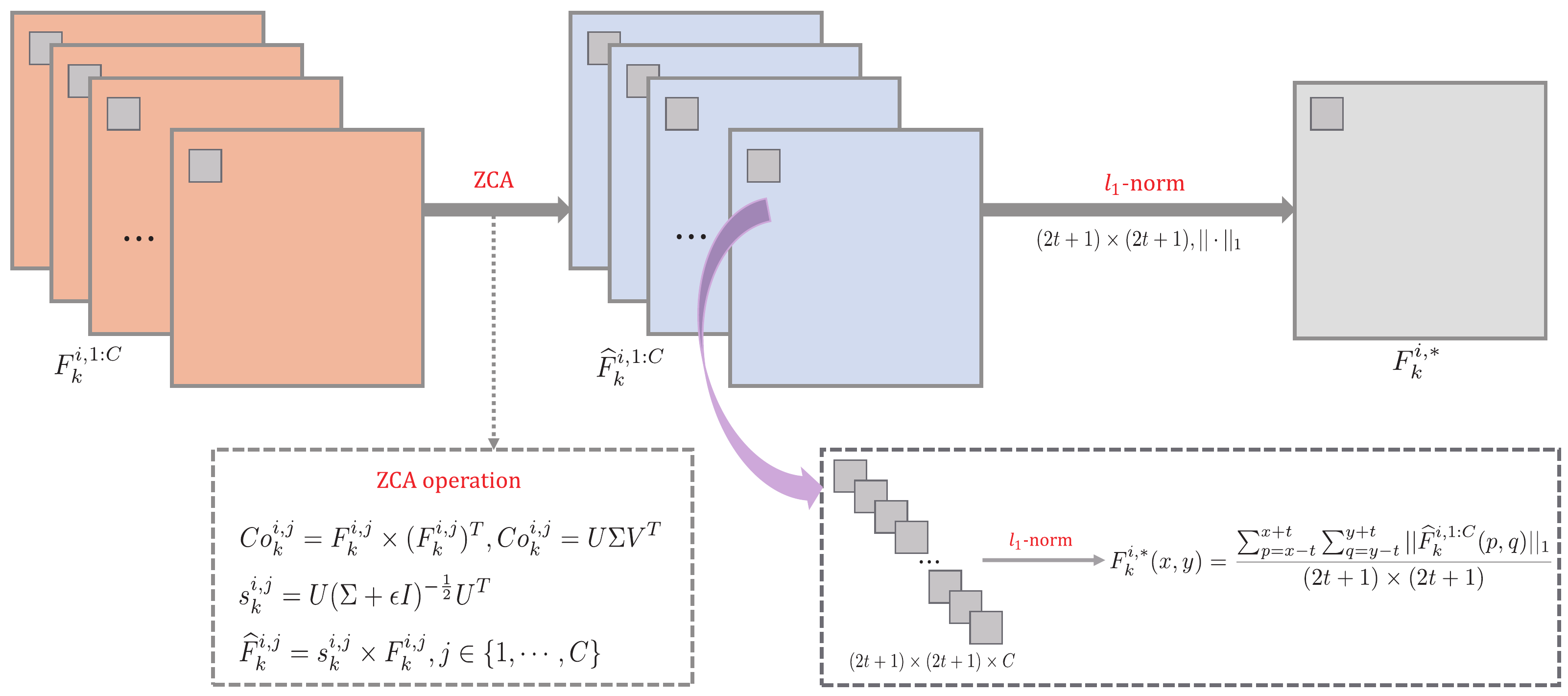}
\caption{The procedure of ZCA and $l_1$-norm operation.}
\label{fig:l1norm}
\end{figure}

After the deep features were generated, we use ZCA to process the deep features $F_k^{i,1:C}$. When we obtain the processed features $\hat{F}_k^{i,1:C}$, we utilize $l_1$-norm to calculate initial weight map $F_k^{i,*}$. The procedure of ZCA and $l_1$-norm operation is shown in Fig.\ref{fig:l1norm}.

$F_k^{i,1:C}$ indicates the deep features obtained by the $i$-th convolutional block, which contains $C$ channels, $k\in{\{1,2\}}$. In ZCA operation, the covariance matrix and its decomposition are calculated by Eq.\ref{eq:3},
\begin{eqnarray}\label{eq:3}
  	&Co_k^{i,j}=F_k^{i,j}\times (F_k^{i,j})^T, \\
	&Co_k^{i,j}=U\Sigma V^T \nonumber
\end{eqnarray}
where $j\in{\{1,2,\cdots,C\}}$ denotes the index of channel in deep features. 

Then we use Eq.\ref{eq:4} to obtain the processed features $\hat{F}_k^{i,1:C}$ which is combined by $\hat{F}_k^{i,j}$,
\begin{eqnarray}\label{eq:4}
  	&\hat{F}_k^{i,j}=s_k^{i,j}\times F_k^{i,j}, \\
	&s_k^{i,j}=U(\Sigma+\epsilon I)^{-\frac{1}{2}}U^T \nonumber
\end{eqnarray}

After we obtain the processed features $\hat{F}_k^{i,1:C}$, we utilize the local $l_1$-norm and average operation to calculate the initial weight maps $F_k^{i,*}$ using Eq.\ref{eq:5},
\begin{eqnarray}\label{eq:5}
  	F_k^{i,*}=\frac{\sum_{p=x-t}^{x+t}\sum_{q=y-t}^{y+t}||\hat{F}_k^{i,1:C}(p,q)||_1}{(2t+1)\times (2t+1)}
\end{eqnarray}

As shown in Fig.\ref{fig:l1norm}, we choose a window which centers at the $\hat{F}_k^{i,1:C}(x,y)$ to calculate the average $l_1$-norm, and in our paper $t=2$.

\subsection{Reconstruction}

\begin{figure}[ht]
\centering
\includegraphics[width=1\linewidth]{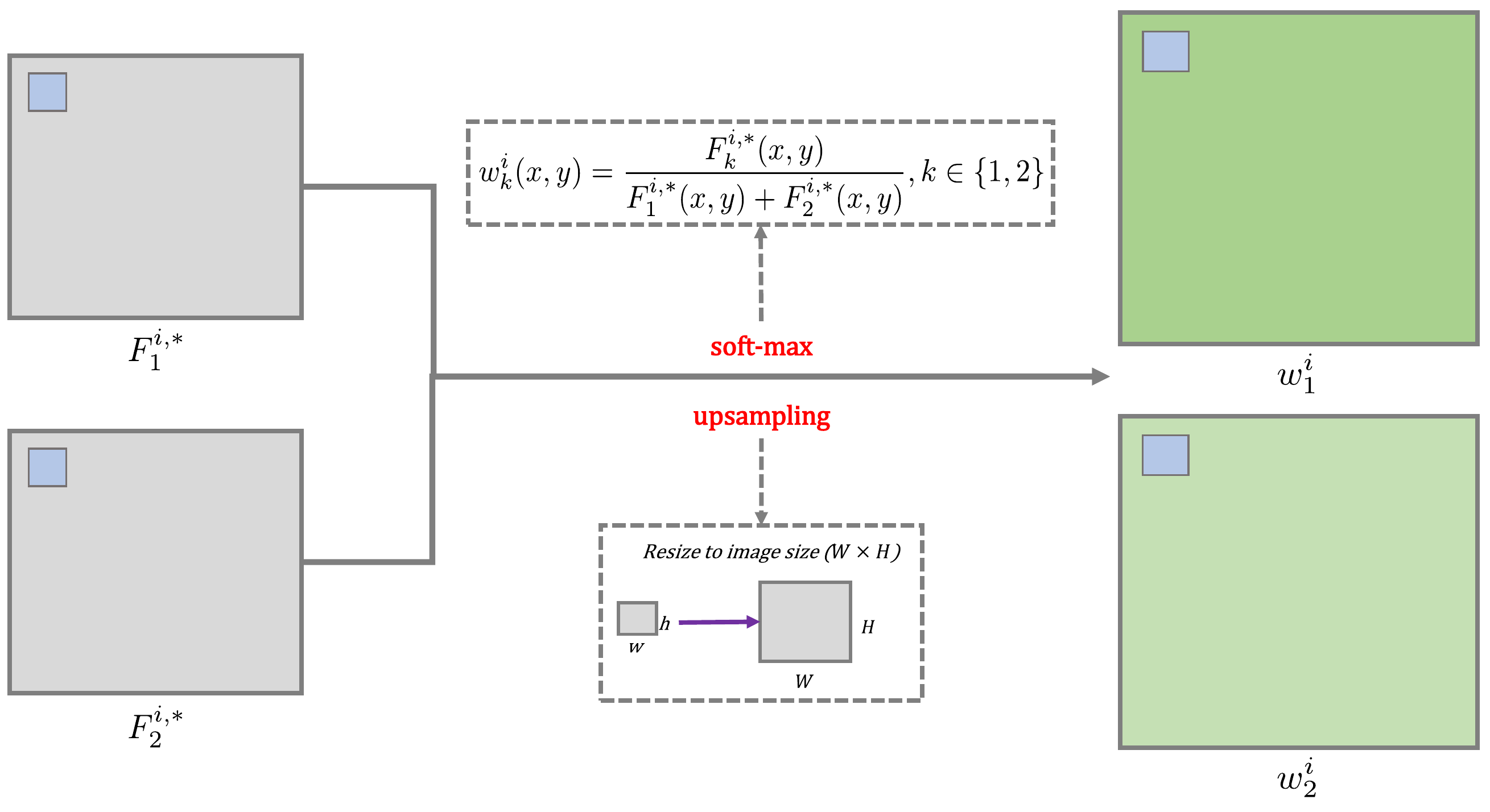}
\caption{Resize and soft-max operation.}
\label{fig:softmfax}
\end{figure}

When the initial weight maps $F_1^{i,*}$ and $F_2^{i,*}$ are calculated by ZCA and $l_1$-norm, the upsampling and soft-max operations are applied to obtain the final weight maps $w_1^i$ and $w_2^i$, as shown in Fig.\ref{fig:softmfax}.

Firstly, the bicubic interpolation which is provided by Matlab is used to resize the initial weight maps into source image size.

Then the final weight maps are obtained by Eq.\ref{eq:6},
\begin{eqnarray}\label{eq:6}
  	w_k^i(x,y)=\frac{F_k^{i,*}(x,y)}{F_1^{i,*}(x,y)+F_2^{i,*}(x,y)}
\end{eqnarray}

Finally, the fused image is reconstructed using Eq.\ref{eq:7},
\begin{eqnarray}\label{eq:7}
  	Fused(x,y)=\sum_{k=1}^2w_k^i(x,y)Source_k(x,y)
\end{eqnarray}

\section{Experiments and Analysis}
\label{experiment}
In this section, the source images and experimental environment are introduced first. Secondly, the effect of different networks and norms in our method are discussed. Then the influence of ZCA operation is analyzed. Finally, the proposed algorithm is evaluated by using subjective and objective criteria. We choose several existing state-of-the-art fusion methods to compare with our algorithm.

\subsection{Experimental Settings}
\begin{figure}[ht]
\centering
\includegraphics[width=\linewidth]{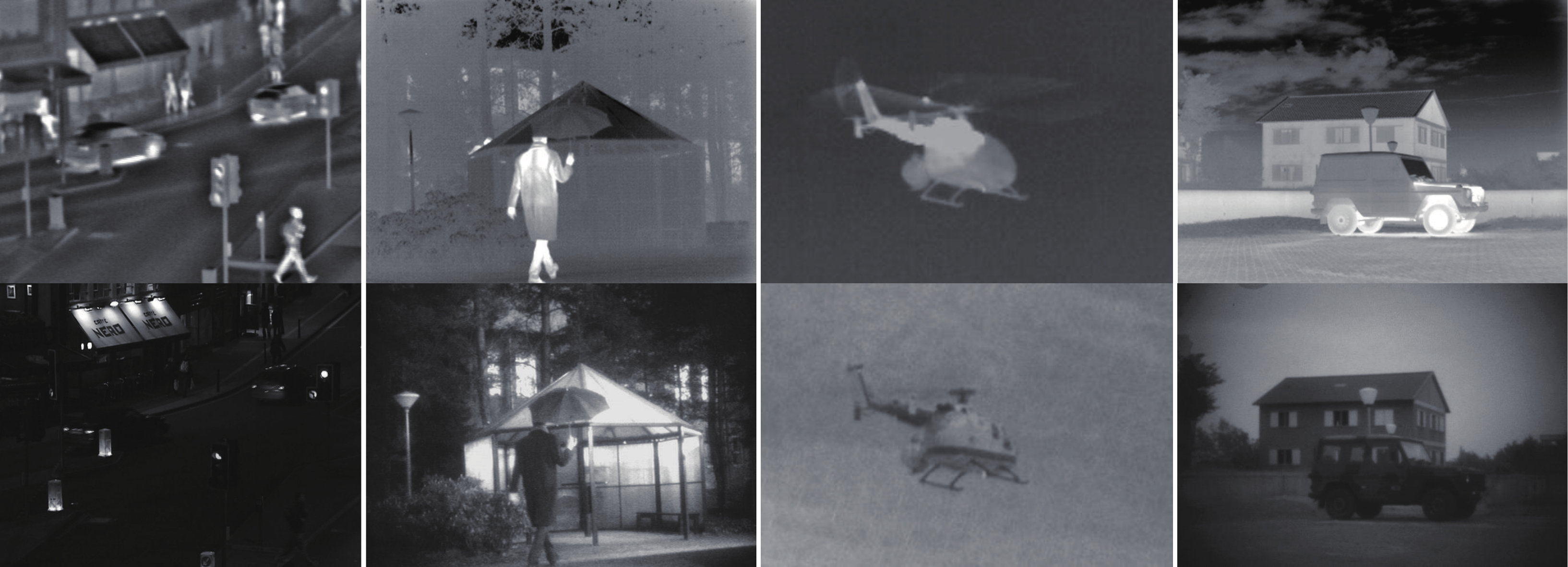}
\caption{Four pairs of source images. The top row contains infrared images, and the second row contains visible images.}
\label{fig:source}
\end{figure}

We collect 21 pairs of source infrared and visible images from \cite{28} and \cite{29}. Our source images are available at \cite{30}. And samples of these source images are shown in Fig.\ref{fig:source}.

In our experiment, DeepFuse\cite{16} is implemented with Tensorflow and GTX 1080Ti, 64 GB RAM. Other fusion algorithms are implemented in MATLAB R2017b on 3.2 GHz Intel(R) Core(TM) CPU with 12 GB RAM. The details of our experiment are introduced in the next sections.

\subsection{The effect of different networks and norms}

In this section, we choose different networks(VGG19\cite{18}, ResNet50\cite{19} and ResNet101\cite{19}) and different norms($l_1$-norm, $l_2$-norm and nuclear-norm\cite{31}) to evaluate the performance of our fusion framework.

\begin{figure}[ht]
\centering
\includegraphics[width=1\linewidth]{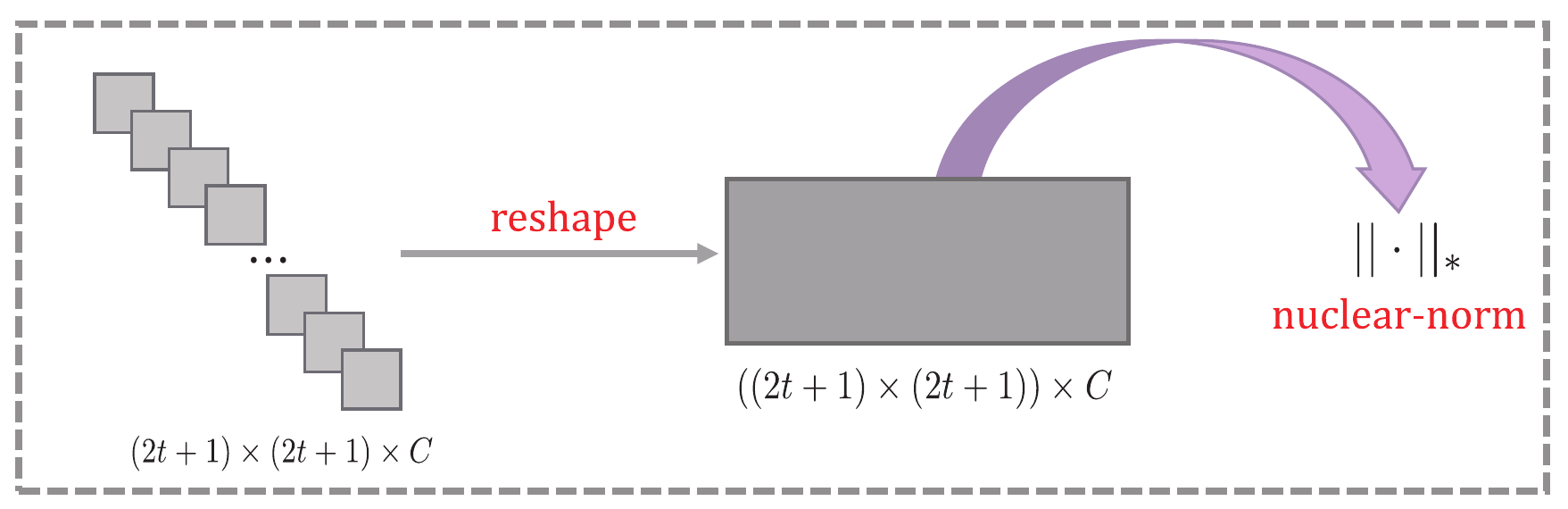}
\caption{The procedure of reshape and nuclear-norm operation.}
\label{fig:nuclearnorm}
\end{figure}

When the nuclear-norm is utilized in our framework, Eq.\ref{eq:5} is rewritten to Eq.\ref{eq:8},
\begin{eqnarray}\label{eq:8}
  	&F_k^{i,*}(x,y)=||R(\hat{F}_k^{i,1:C}[(x-t):(x+t),(y-t):(y+t)])||_*, \\
	&s.t., t=2 \nonumber
\end{eqnarray}
where $R(\cdot)$ indicates the reshape operation and $R(\cdot)\in{\mathbb{R}^{[(2t+1)\times (2t+1)]\times C}}$. The reshape and nuclear-norm operation are shown in Fig.\ref{fig:nuclearnorm}.

Five quality metrics are utilized to assess the performance. These are: $FMI_{pixel}$\cite{32}, $FMI_{dct}$\cite{32} and $FMI_w$\cite{32} which calculate mutual information (FMI) for the pixel, discrete cosine and wavelet features, respectively; $N_{abf}$\cite{33} denotes the rate of noise or artifacts added to the fused image by the fusion process; and modified structural similarity($SSIM_a$)\cite{17}.

The performance improves with the increasing numerical index of$FMI_{pixel}$, $FMI_{dct}$, $FMI_w$ and $SSIM_a$. Also, the fusion performance is better when the value of $N_{abf}$ is small which means the fused images contain less artificial information and noise.

We calculate the average quality metrics values of 21 pairs of source images. In VGG19, the outputs of four layers(relu1\_1, relu2\_1, relu3\_1, relu4\_1) are used. In ResNet50 and ResNet101, we choose four convolutional blocks (Conv2, Conv3, Conv4, Conv5). These values are shown in Table \ref{table:1} \ref{table:2} \ref{table:3}.
\begin{table}[ht]
\centering
\caption{\label{table:1}Quality metrics values - Our fusion framework use $l_1$-norm and \textbf{different networks}.}
\resizebox{\textwidth}{!}{
\begin{tabular}{c |c| c c c c c}
\hline
\multicolumn{7}{c}{$l_1$-norm} \\
\hline
\multicolumn{2}{c|}{Networks} & $FMI_{pixel}$ & $FMI_{dct}$ & $FMI_w$ & $N_{abf}$ & $SSIM_a$ \\
\hline
 \multirow{4}*{VGG19} &
 relu1\_1	&0.90120	&0.36208	&0.35872	&0.11696	&0.73833\\
&relu2\_1	&0.91030	&0.39170	&0.40032	&0.04227	&0.76469\\
&relu3\_1	&0.91122	&0.39399	&0.40948	&0.01309	&0.77326\\
&relu4\_1	&0.91057	&0.39666	&0.41306	&0.00397	&0.77613\\
\hline
\multirow{4}*{ResNet50} &
 Conv2		&\textbf{0.91257}	&0.39545	&0.41126	&0.01495	&0.77251\\
&Conv3		&0.91156	&0.39651	&0.41442	&0.00468	&0.77561\\
&Conv4		&0.91093	&0.40296	&0.41652	&0.00131	&0.77749\\
&Conv5		&0.90921	&\color{red}{0.40577}	&\color{red}{0.41689}	&\textbf{0.00062}	&\textbf{0.77825}\\
\hline
\multirow{4}*{ResNet101} &
 Conv2		&\color{red}{0.91255}	&0.39472	&0.41089	&0.01599	&0.77215\\
&Conv3		&0.91135	&0.39589	&0.41383	&0.00510	&0.77544\\
&Conv4		&0.90961	&0.40386	&0.41643	&\color{red}{0.00091}	&0.77791\\
&Conv5		&0.90934	&\textbf{0.40605}	&\textbf{0.41706}	&\textbf{0.00062}	&\color{red}{0.77821}\\
\hline
\end{tabular}}
\end{table}

\begin{table}[p]
\centering
\caption{\label{table:2}Quality metrics values - Our fusion framework use $l_2$-norm and \textbf{different networks}.}
\resizebox{\textwidth}{!}{
\begin{tabular}{c |c| c c c c c}
\hline
\multicolumn{7}{c}{$l_2$-norm} \\
\hline
\multicolumn{2}{c|}{Networks} & $FMI_{pixel}$ & $FMI_{dct}$ & $FMI_w$ & $N_{abf}$ & $SSIM_a$ \\
\hline
\multirow{4}*{VGG19} &
 relu1\_1	&0.90191 	&0.36500 	&0.36172 	&0.11063 	&0.74113 \\
&relu2\_1	&0.91042 	&0.39217 	&0.40078 	&0.04053 	&0.76529 \\
&relu3\_1	&0.91118 	&0.39433 	&0.40962 	&0.01272 	&0.77344 \\
&relu4\_1	&0.91054 	&0.39696 	&0.41312 	&0.00381 	&0.77622 \\
\hline
\multirow{4}*{ResNet50} &
 Conv2	&\textbf{0.91255} 	&0.39522 	&0.41088 	&0.01487 	&0.77263 \\
&Conv3	&0.91146 	&0.39635 	&0.41430 	&0.00470 	&0.77562 \\
&Conv4	&0.91087 	&0.40265 	&0.41645 	&0.00134 	&0.77746 \\
&Conv5	&0.90925 	&\color{red}{0.40544} 	&\color{red}{0.41654} 	&\textbf{0.00064} 	&\textbf{0.77825} \\
\hline
\multirow{4}*{ResNet101} &
 Conv2	&\color{red}{0.91254} 	&0.39468 	&0.41059 	&0.01576 	&0.77230 \\
&Conv3	&0.91126 	&0.39584 	&0.41366 	&0.00511 	&0.77547 \\
&Conv4	&0.90965 	&0.40385 	&0.41643 	&\color{red}{0.00091} 	&\color{red}{0.77792} \\
&Conv5	&0.90932 	&\textbf{0.40561} 	&\textbf{0.41666} 	&\textbf{0.00064} 	&\textbf{0.77825} \\
\hline
\end{tabular}}
\end{table}

\begin{table}[p]
\centering
\caption{\label{table:3}Quality metrics values - Our fusion framework use nuclear-norm and \textbf{different networks}.}
\resizebox{\textwidth}{!}{
\begin{tabular}{c |c| c c c c c}
\hline
\multicolumn{7}{c}{nuclear-norm} \\
\hline
\multicolumn{2}{c|}{Networks} & $FMI_{pixel}$ & $FMI_{dct}$ & $FMI_w$ & $N_{abf}$ & $SSIM_a$ \\
\hline
\multirow{4}*{VGG19} &
 relu1\_1	&0.90505 	&0.37650 	&0.37536 	&0.07989 	&0.75391 \\
&relu2\_1	&0.91040 	&0.39454 	&0.40288 	&0.03176 	&0.76845 \\
&relu3\_1	&0.91092 	&0.39546 	&0.41017 	&0.01125 	&0.77403 \\
&relu4\_1	&0.91045 	&0.39758 	&0.41338 	&0.00349 	&0.77637 \\
\hline
\multirow{4}*{ResNet50} &
 Conv2		&\textbf{0.91274} 	&0.39659 	&0.41178 	&0.01320 	&0.77329 \\
&Conv3		&0.91177 	&0.39712 	&0.41483 	&0.00439 	&0.77566 \\
&Conv4		&0.91110 	&0.40238 	&\textbf{0.41673} 	&0.00141 	&0.77726 \\
&Conv5		&0.90932 	&\color{red}{0.40509} 	&0.41648 	&\color{red}{0.00067} 	&\color{red}{0.77817} \\
\hline
\multirow{4}*{ResNet101} &
 Conv2		&\color{red}{0.91266} 	&0.39621 	&0.41166 	&0.01359 	&0.77304 \\
&Conv3		&0.91148 	&0.39651 	&0.41410 	&0.00470 	&0.77557 \\
&Conv4		&0.90971 	&0.40351 	&0.41641 	&0.00095 	&0.77784 \\
&Conv5		&0.90943 	&\textbf{0.40537} 	&\color{red}{0.41655} 	&\textbf{0.00065} 	&\textbf{0.77820} \\
\hline
\end{tabular}}
\end{table}

The best values are indicated in bold, the second best values are indicated in red font. As we can see, the ResNets(50/101) obtain all the best and the second best values in different norms. This means ResNet can achieve better fusion performance than VGG19 in our fusion framework.
 
Comparing ResNet50 with ResNet101 in Table \ref{table:1} \ref{table:2} \ref{table:3}, the quality metrics values are very close. Considering the time efficiency, in our method, the ResNet50 is utilized.

\begin{table}[ht]
\centering
\caption{\label{table:4}Quality metrics values - Our fusion framework use ResNet50 and \textbf{different norms}.}
\resizebox{\textwidth}{!}{
\begin{tabular}{c |c| c c c c c}
\hline
\multicolumn{7}{c}{ResNet50} \\
\hline
\multicolumn{2}{c|}{Networks} & $FMI_{pixel}$ & $FMI_{dct}$ & $FMI_w$ & $N_{abf}$ & $SSIM_a$ \\
\hline
\multirow{4}*{$l_1$-norm} &
 Conv2	&\color{red}{0.91257} 	&0.39545 	&0.41126 	&0.01495 	&0.77251 \\
&Conv3	&0.91156 	&0.39651 	&0.41442 	&0.00468 	&0.77561 \\
&Conv4	&0.91093 	&0.40296 	&0.41652 	&0.00131 	&0.77749 \\
&Conv5	&0.90921 	&\textbf{0.40577} 	&\textbf{0.41689} 	&\textbf{0.00062} 	&\textbf{0.77825} \\
\hline
\multirow{4}*{$l_2$-norm} &
 Conv2	&0.91255 	&0.39522 	&0.41088 	&0.01487 	&0.77263 \\
&Conv3	&0.91146 	&0.39635 	&0.41430 	&0.00470 	&0.77562 \\
&Conv4	&0.91087 	&0.40265 	&0.41645 	&0.00134 	&0.77746 \\
&Conv5	&0.90925 	&\color{red}{0.40544} 	&0.41654 	&\color{red}{0.00064} 	&\textbf{0.77825} \\
\hline
\multirow{4}*{nuclear-norm} &
 Conv2	&\textbf{0.91274} 	&0.39659 	&0.41178 	&0.01320 	&0.77329 \\
&Conv3	&0.91177 	&0.39712 	&0.41483 	&0.00439 	&0.77566 \\
&Conv4	&0.91110 	&0.40238 	&\color{red}{0.41673} 	&0.00141 	&0.77726 \\
&Conv5	&0.90932 	&0.40509 	&0.41648 	&0.00067 	&\color{red}{0.77817} \\
\hline
\end{tabular}}
\end{table}

In Table \ref{table:4}, we evaluate the effect of different norms with ResNet50 in our method. From Table \ref{table:4}, $l_1$-norm contains four best values and one second best values. This means, in our fusion framework, $l_1$-norm has better performance than other norms.

\subsection{The influence of ZCA operation}

In this section, we analyze the influence of ZCA operation on our method. We choose ResNet50 and three norms($l_1$-norm, $l_2$-norm and nuclear-norm[31]) to evaluate the performance with or without ZCA.

Ten quality metrics are chosen. These metrics include: En(entropy), MI(mutual information), $Q_{abf}$\cite{34}, $FMI_{pixel}$\cite{32}, $FMI_{dct}$\cite{32}, $FMI_w$\cite{32}, $N_{abf}$\cite{33}, SCD\cite{35}, $SSIM_a$\cite{17}, and MS\_SSIM\cite{36}. The performance improves with the increasing numerical index of En, MI, $Q_{abf}$, $FMI_{pixel}$, $FMI_{dct}$, $FMI_w$, SCD, $SSIM_a$ and MS\_SSIM. However, its better when the values of $N_{abf}$ are small.

\begin{table}[ht]
\centering
\caption{\label{table:5}Quality metrics values - Our fusion framework use ResNet50 and \textbf{ZCA operation}.}
\resizebox{\textwidth}{!}{
\begin{tabular}{c |c| c c c c c c c c c c}
\hline
\multicolumn{2}{c|}{norms} &En &MI &$Q_{abf}$ & $FMI_{pixel}$ & $FMI_{dct}$ & $FMI_w$ & $N_{abf}$ &SCD &$SSIM_a$ &MS\_SSIM \\
\hline
\multirow{4}*{$l_1$-norm} &
 Conv2	&\textbf{6.29026}	&\textbf{12.58052}	&\textbf{0.40154}	&0.91257	&0.39545	&0.41126	&0.01495	&1.64113	&0.77251	&0.87204\\
&Conv3	&6.28155	&12.56309	&0.39314	&0.91156	&0.39651	&0.41442	&0.00468	&\textbf{1.64235}	&0.77561	&\textbf{0.88102}\\
&Conv4	&6.23540	&12.47081	&0.37254	&0.91093	&0.40296	&0.41652	&0.00131	&1.63949	&0.77749	&0.87962\\
&Conv5	&6.19527	&12.39054	&0.35098	&0.90921	&\textbf{0.40577}	&\textbf{0.41689}	&\textbf{0.00062}	&1.63358	&\textbf{0.77825}	&0.87324\\
\hline
\multirow{4}*{$l_2$-norm} &
 Conv2	&6.28650	&12.57299	&0.39898	&0.91255	&0.39522	&0.41088	&0.01487	&1.63984	&0.77263	&0.87103\\
&Conv3	&6.28027	&12.56054	&0.39215	&0.91146	&0.39635	&0.41430	&0.00470	&\color{red}{1.64175}	&0.77562	&0.88049\\
&Conv4	&6.23730	&12.47459	&0.37283	&0.91087	&0.40265	&0.41645	&0.00134	&1.63951	&0.77746	&0.87954\\
&Conv5	&6.19689	&12.39377	&0.35080	&0.90925	&\color{red}{0.40544}	&0.41654	&\color{red}{0.00064}	&1.63394	&\textbf{0.77825}	&0.87312\\
\hline
\multirow{4}*{nuclear-norm} &
 Conv2	&6.28192	&12.56384	&\color{red}{0.39986}	&\textbf{0.91274}	&0.39659	&0.41178	&0.01320	&1.63936	&0.77329	&0.87286\\
&Conv3	&\color{red}{6.28654}	&\color{red}{12.57309}	&0.39473	&0.91177	&0.39712	&0.41483	&0.00439	&1.64087	&0.77566	&\color{red}{0.88094}\\
&Conv4	&6.25057	&12.50114	&0.37713	&0.91110	&0.40238	&\color{red}{0.41673}	&0.00141	&1.63960	&0.77726	&0.88047\\
&Conv5	&6.20433	&12.40865	&0.35311	&0.90932	&0.40509	&0.41648	&0.00067	&1.63431	&\color{red}{0.77817}	&0.87374\\
\hline
\end{tabular}}
\end{table}

\begin{table}[ht]
\centering
\caption{\label{table:6}Quality metrics values - Our fusion framework use ResNet50 but \textbf{without ZCA operation}.}
\resizebox{\textwidth}{!}{
\begin{tabular}{c |c| c c c c c c c c c c}
\hline
\multicolumn{2}{c|}{norms} &En &MI &$Q_{abf}$ & $FMI_{pixel}$ & $FMI_{dct}$ & $FMI_w$ & $N_{abf}$ &SCD &$SSIM_a$ &MS\_SSIM \\
\hline
\multirow{4}*{$l_1$-norm} &
 Conv2	&6.17245 	&12.34490 	&0.34191 	&0.90884 	&0.40631 	&0.41678 	&0.00060 	&1.62953 	&\color{red}{0.77848} 	&0.87014 \\
&Conv3	&6.17108 	&12.34216 	&0.34168 	&0.90866 	&0.40652 	&0.41701 	&0.00058 	&1.62854 	&0.77847 	&0.86995 \\
&Conv4	&6.17751 	&12.35501 	&0.34468 	&\color{red}{0.90908} 	&0.40690 	&0.41739 	&0.00051 	&\color{red}{1.63178} 	&0.77844 	&\color{red}{0.87154} \\
&Conv5	&\color{red}{6.17760} 	&\color{red}{12.35519} 	&\color{red}{0.34506} 	&0.90885 	&0.40669 	&0.41737 	&0.00052 	&1.63037 	&0.77844 	&0.87125 \\
\hline
\multirow{4}*{$l_2$-norm} &
 Conv2	&6.17225 	&12.34451 	&0.34157 	&0.90883 	&0.40647 	&0.41697 	&0.00057 	&1.62945 	&\textbf{0.77850} 	&0.87021 \\
&Conv3	&6.17078 	&12.34157 	&0.34125 	&0.90864 	&0.40641 	&0.41697 	&0.00057 	&1.62824 	&0.77847 	&0.86982 \\
&Conv4	&6.17612 	&12.35224 	&0.34361 	&0.90898 	&0.40671 	&0.41729 	&\color{red}{0.00052} 	&1.63121 	&0.77846 	&0.87110 \\
&Conv5	&6.17539 	&12.35078 	&0.34462 	&0.90877 	&\color{red}{0.40692} 	&\textbf{0.41751} 	&\textbf{0.00049} 	&1.62996 	&0.77845 	&0.87126 \\
\hline
\multirow{4}*{nuclear-norm} &
 Conv2	&\textbf{6.18154} 	&\textbf{12.36308} 	&\textbf{0.35226} 	&\textbf{0.90969} 	&\textbf{0.40701} 	&0.41731 	&0.00064 	&\textbf{1.63449} 	&0.77832 	&\textbf{0.87363} \\
&Conv3	&\textbf{6.19141} 	&\textbf{12.38282} 	&\textbf{0.35886} 	&\textbf{0.91029} 	&0.40645 	&0.41721 	&0.00067 	&\textbf{1.63843} 	&0.77823 	&\textbf{0.87651} \\
&Conv4	&\textbf{6.18871} 	&\textbf{12.37742} 	&\textbf{0.35321} 	&\textbf{0.90978} 	&0.40635 	&0.41721 	&0.00057 	&\textbf{1.63578} 	&0.77830 	&\textbf{0.87460} \\
&Conv5	&\textbf{6.18116} 	&\textbf{12.36232} 	&\textbf{0.34817} 	&\textbf{0.90909} 	&0.40679 	&\color{red}{0.41746} 	&\color{red}{0.00052} 	&\textbf{1.63205} 	&0.77840 	&\textbf{0.87243} \\
\hline
\end{tabular}}
\end{table}

Table \ref{table:5} and Table \ref{table:6} show the quality values with and without ZCA, respectively. In Table \ref{table:6}, when the ZCA is not used, ResNet50 with nuclear-norm achieves the best values. This means the low-rank ability is more useful than other norms in original deep features. However, in Table \ref{table:5}, when we use ZCA to project deep features into a sub-space, $l_1$-norm will obtain most of the best values, even compared with nuclear-norm + without ZCA. We think the ZCA projects the original data into a sparse space, and in this situation, the sparse metric($l_1$-norm) obtains better performance than low-rank metric(nuclear-norm).

Based on above observation, we choose ResNet50 to extract deep features in our fusion method, ZCA and $l_1$-norm operations are used to obtain initial weight maps.

\subsection{Subjective Evaluation}

In subjective and objective evaluation, we choose nine existing fusion methods to compare with our algorithm. These fusion methods are: cross bilateral filter fusion method(CBF)\cite{37}, discrete cosine harmonic wavelet transform(DCHWT)\cite{33}, joint sparse representation(JSR)\cite{8}, saliency detection in sparse domain(JSRSD)\cite{38}, gradient transfer and total variation minimization(GTF)\cite{39}, weighted least square optimization(WLS)\cite{28}, convolutional sparse representation(ConvSR)\cite{14}, a Deep Learning Framework based on VGG19 and multi-layers(VggML)\cite{17}, and DeepFuse\cite{16}.

\begin{figure}[ht]
\centering
\includegraphics[width=1\linewidth]{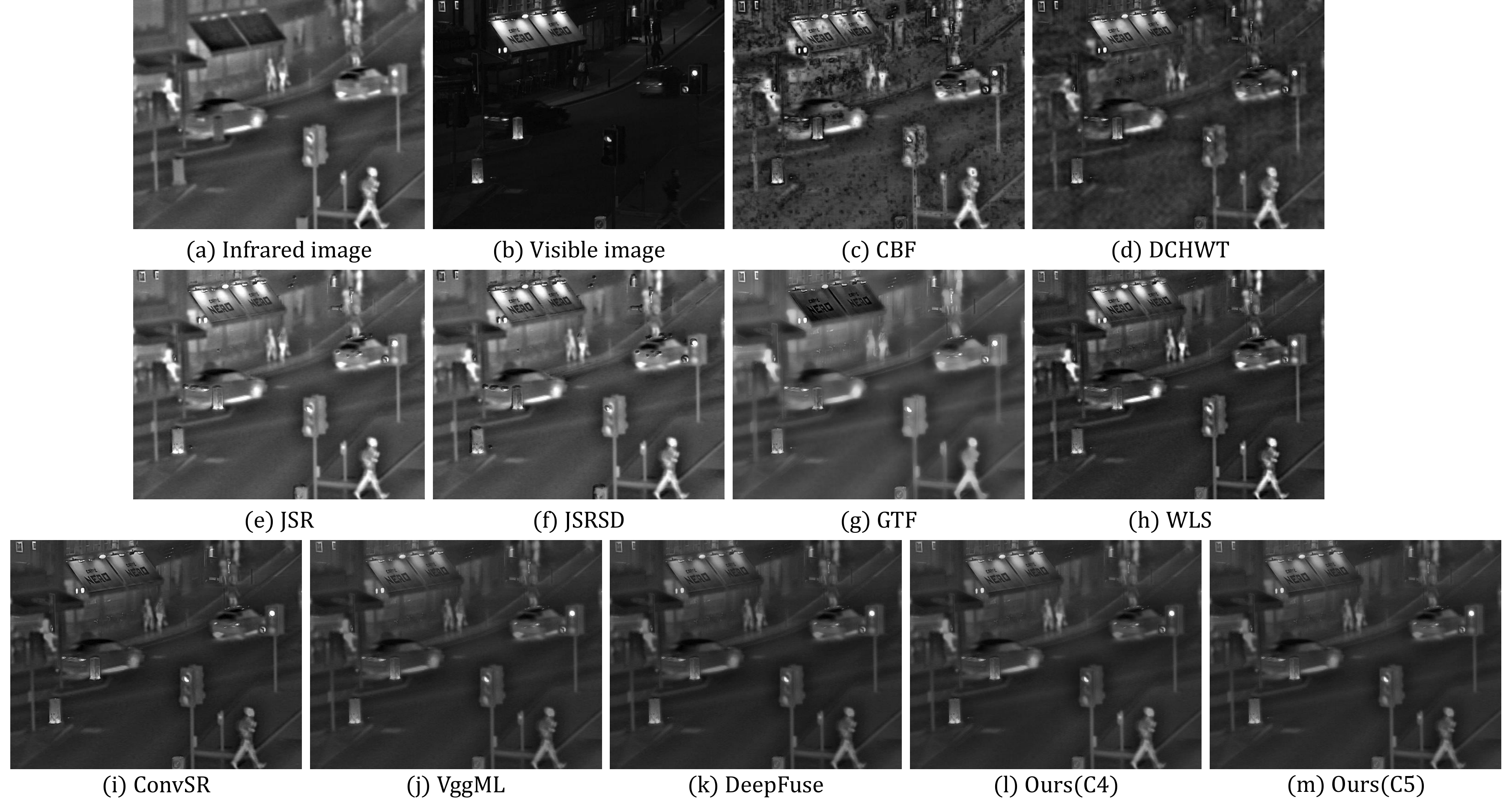}
\caption{Experiment on ``street'' images. (a) Infrared image; (b) Visible image; (c) CBF; (d) DCHWT; (e) JSR; (f) JSRSD. (g) GTF; (h) WLS; (i)ConvSR; (j)VggML; (k)DeepFuse; (l)ours(Conv4); (m) ours(Conv5).}
\label{fig:result}
\end{figure}

In our fusion method, the convolutional blocks (Conv4 and Conv5) are chosen to obtain the fused images. The fused images are shown in Fig.\ref{fig:result}. As an example, we evaluate the relative performance of the fusion methods only on a single pair of images (``street'').

From Fig.\ref{fig:result}(c-m), the fused images which are obtained by CBF and DCHWT contain more noise and some saliency features are not clear. The JSR, JSRSD, GTF and WLS can obtain better performance and less noise. But these fused images still contain artificial information near the saliency features. On the contrary, deep learning-based fusion methods, such as ConvSR, VggML, DeepFuse and ours, contain more saliency features and preserve more detail information, and the fused images look more natural. As there is no validation difference between these deep learning-based methods and the proposed algorithm in terms of human sensitivity, we choose several objective quality metrics to assess the fusion performance in the next section.

\subsection{Objective Evaluation}

For the purpose of quantitative comparison between the proposed method and existing fusion methods, four quality metrics are utilized. These are: $FMI_{pixel}$\cite{32}, $N_{abf}$\cite{33}, $SSIM_a$\cite{17} and average Edge Preservation Index($EPI_a$).

\begin{figure}[ht]
\centering
\includegraphics[width=1\linewidth]{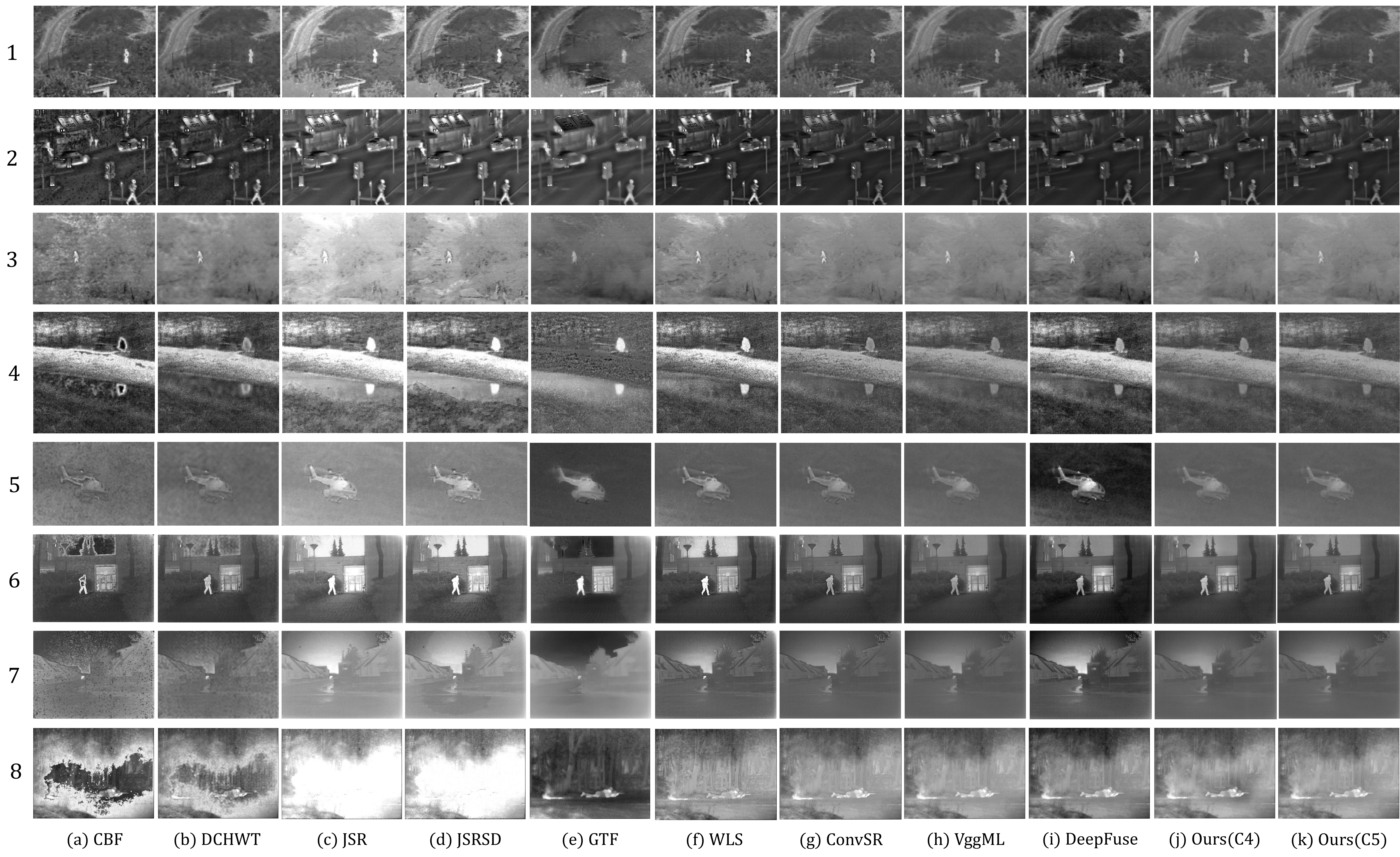}
\caption{Experiment on 8 pairs images. (a)CBF; (b)DCHWT; (c)JSR; (d)JSRSD; (e)GTF; (f)WLS; (g)ConvSR; (h)VggML; (i)DeepFuse; (j)Ours(Conv4); (k)ours(Conv5);}
\label{fig:results8}
\end{figure}

The average EPI($EPI_a$) is calculated by Eq. \ref{eq:9},
\begin{eqnarray}\label{eq:9}
  	EPI_a(F,I_1,I_2)=[EPI(F,I_1)+EPI(F,I_2)]\times 0.5
\end{eqnarray}
where $EPI(\cdot)$ indicates the edge preservation index operation \cite{40}.

In this section, we choose 8 pairs of images to evaluate the performance of the existing methods and our fusion algorithm. The fused images are shown in Fig.\ref{fig:results8}, and the values of $FMI_{pixel}$, $N_{abf}$, $SSIM_a$ and $EPI_a$ are presented in Table \ref{table:7}. The best values are indicated in bold.

\begin{table}[ht]
\centering
\caption{\label{table:7}The values of $FMI_{pixel}$\cite{32}, $N_{abf}$\cite{33}, $SSIM_a$\cite{17} and $EPI_a$\cite{40} for 8 pairs images.}
\resizebox{\textwidth}{!}{
\begin{tabular}{c |c| c c c c c c c c c c c}
\hline
\multirow{2}*{Images}	&\multirow{2}*{Metrics}	&\multirow{2}*{CBF}	&\multirow{2}*{DCHWT}	&\multirow{2}*{JSR} 	&\multirow{2}*{JSRSD}	&\multirow{2}*{GTF} 	&\multirow{2}*{WLS}	&\multirow{2}*{ConvSR}	 &\multirow{2}*{VggML}	 &\multirow{2}*{DeepFuse} & \multicolumn{2}{c}{Ours}\\
\cline{12-13}
 & & & & & & & & & & &Conv4	&Conv5\\
\hline
\multirow{3}*{Fig.\ref{fig:results8} (1)} &
$FMI_{pixel}$	&0.87010	&0.89000	&0.85281	&0.83392	&0.88393	&0.87897	&\textbf{0.89724}	&0.88532	&0.88226	&0.88517	&0.88229\\
&$N_{abf}$	&0.23167	&0.05118	&0.34153	&0.34153	&0.07027	&0.14494	&0.01494	&0.00013	&0.03697	&\textbf{0.00011}	&\textbf{0.00008}\\
&$SSIM_a$		&0.62376	&0.74834	&0.52715	&0.52715	&0.69181	&0.72827	&0.74954	&0.77758	&0.73314	&\textbf{0.77765}	&\textbf{0.77773}\\
&$EPI_a$		&0.68093 	&0.78324 	&0.68074 	&0.52103 	&0.72270 	&0.74477 	&0.77789 	&0.83152 	&0.79487 	&\textbf{0.83194} 	&\textbf{0.83218}\\
\hline
\multirow{3}*{Fig.\ref{fig:results8} (2)} &
$FMI_{pixel}$	&0.89441	&0.92106	&0.91004	&0.90446	&0.91526	&0.91144	&\textbf{0.92269}	&0.91849	&0.91763	&0.92068	&0.91851\\
&$N_{abf}$	&0.48700	&0.21840	&0.19749	&0.19889	&0.11237	&0.16997	&0.02199	&0.00376	&0.00262	&\textbf{0.00367}	&\textbf{0.00222}\\
&$SSIM_a$		&0.49861	&0.64468	&0.62399	&0.62353	&0.61109	&0.66873	&0.67474	&0.68041	&0.68092	&\textbf{0.68130}	&\textbf{0.68039}\\
&$EPI_a$		&0.60258 	&0.70134 	&0.66876 	&0.64165 	&0.58663 	&0.66523 	&0.69587 	&0.72348 	&0.72569 	&0.72344 	&\textbf{0.72725}\\
\hline
\multirow{3}*{Fig.\ref{fig:results8} (3)} &
$FMI_{pixel}$	&0.80863	&0.86116	&0.78628	&0.77685	&0.83492	&0.85603	&\textbf{0.87225}	&0.85276	&0.84882	&0.85248	&0.84916\\
&$N_{abf}$	&0.43257	&0.07415	&0.49804	&0.49804	&0.08501	&0.19188	&0.00991	&0.00020	&0.09275	&\textbf{0.00013}	&\textbf{0.00009}\\
&$SSIM_a$		&0.59632	&0.80619	&0.46767	&0.46767	&0.73386	&0.77506	&0.81383	&0.84569	&0.81415	&\textbf{0.84607}	&\textbf{0.84622}\\
&$EPI_a$		&0.74569 	&0.87767 	&0.76349 	&0.60302 	&0.77089 	&0.83996 	&0.88052 	&\textbf{0.90449} 	&0.86859 	&0.90445 	&0.90443 \\
\hline
\multirow{3}*{Fig.\ref{fig:results8} (4)} &
$FMI_{pixel}$	&0.85685	&\textbf{0.87010}	&0.84809	&0.84340	&0.83589	&0.84038	&0.86855	&0.86309	&0.86206	&0.86329	&0.86198\\
&$N_{abf}$	&0.15233	&0.05781	&0.21640	&0.21536	&0.12329	&0.23343	&0.03404	&0.00037	&0.11997	&\textbf{0.00007}	&\textbf{0.00004}\\
&$SSIM_a$		&0.52360	&0.57614	&0.45422	&0.45458	&0.50273	&0.55427	&0.56129	&0.61117	&0.59249	&\textbf{0.61280}	&\textbf{0.61306}\\
&$EPI_a$		&0.59838 	&0.64235 	&0.61766 	&0.54948 	&0.64006 	&0.64084 	&0.64205 	&0.72925 	&0.67661 	&\textbf{0.73698} 	&\textbf{0.73742}\\
\hline
\multirow{3}*{Fig.\ref{fig:results8} (5)} &
$FMI_{pixel}$	&0.89101	&0.92772	&0.90630	&0.88746	&0.93516	&0.90851	&\textbf{0.94036}	&0.93248	&0.93174	&0.93244	&0.93154\\
&$N_{abf}$	&0.47632	&0.10340	&0.33225	&0.32941	&0.07322	&0.20588	&0.01022	&0.00109	&0.50900	&\textbf{0.00044}	&\textbf{0.00033}\\
&$SSIM_a$		&0.66486	&0.83815	&0.70211	&0.70365	&0.80499	&0.82727	&0.85822	&0.87250	&0.59381	&\textbf{0.87303}	&\textbf{0.87310}\\
&$EPI_a$		&0.81066 	&0.92462 	&0.86499 	&0.75362 	&0.82711 	&0.87144 	&0.92110 	&0.94188 	&0.68235 	&\textbf{0.94378} 	&\textbf{0.94359}\\
\hline
\multirow{3}*{Fig.\ref{fig:results8} (6)} &
$FMI_{pixel}$	&0.89679	&0.93556	&0.91893	&0.89321	&0.93693	&0.92803	&\textbf{0.94206}	&0.93491	&0.93349	&0.93889	&0.93481\\
&$N_{abf}$	&0.25544	&0.07260	&0.32488	&0.32502	&0.03647	&0.22335	&0.01545	&0.00058	&0.00948	&\textbf{0.00055}	&\textbf{0.00023}\\
&$SSIM_a$		&0.64975	&0.75453	&0.58298	&0.58333	&0.70077	&0.72693	&0.76111	&0.78692	&0.72572	&\textbf{0.78709}	&\textbf{0.78743}\\
&$EPI_a$		&0.59079 	&0.74987 	&0.62528 	&0.49270 	&0.66881 	&0.71674 	&0.72890 	&0.79187 	&0.77024 	&0.79160 	&\textbf{0.79393}\\
\hline
\multirow{3}*{Fig.\ref{fig:results8} (7)} &
$FMI_{pixel}$	&0.82883	&0.92680	&0.91361	&0.89055	&0.92438	&0.90933	&\textbf{0.93853}	&0.93161	&0.93066	&0.93309	&0.93024\\
&$N_{abf}$	&0.52887	&0.19714	&0.33544	&0.33720	&0.03276	&0.31160	&0.01561	&0.00122	&0.29958	&\textbf{0.00120}	&\textbf{0.00069}\\
&$SSIM_a$		&0.50982	&0.72735	&0.60153	&0.60078	&0.69419	&0.72919	&0.77048	&0.78256	&0.73096	&0.78201	&\textbf{0.78285}\\
&$EPI_a$		&0.54331 	&0.75715 	&0.72762 	&0.64584 	&0.74966 	&0.73843 	&0.75911 	&0.81527 	&0.74404 	&0.81748 	&\textbf{0.81935}\\
\hline
\multirow{3}*{Fig.\ref{fig:results8} (8)} &
$FMI_{pixel}$	&0.87393	&\textbf{0.94915}	&0.93052	&0.90994	&0.91430	&0.92903	&0.94804	&0.94660	&0.92682	&0.94501	&0.94443\\
&$N_{abf}$	&0.25892	&0.24507	&0.16588	&0.16541	&0.09293	&0.18401	&0.02574	&0.00203	&0.00175	&0.00706	&\textbf{0.00130}\\
&$SSIM_a$		&0.53005	&0.62304	&0.57422	&0.57412	&0.62966	&0.67908	&0.70304	&0.72860	&0.71540	&0.72708	&\textbf{0.72864}\\
&$EPI_a$		&0.31405 	&0.51700 	&0.37213 	&0.28576 	&0.52455 	&0.51057 	&0.53468 	&0.63151 	&0.61975 	&\textbf{0.63973} 	&\textbf{0.64403}\\
\hline
\end{tabular}}
\end{table}

From Fig.\ref{fig:results8} and Table \ref{table:7}, our method achieves better fusion performance in subjective and objective evaluation. In Table \ref{table:7}, the best values are indicated in bold. Compared with other existing fusion methods, our algorithm obtains almost all the best values in $N_{abf}$, $SSIM_a$ and $EPI_a$, which represent that the fused images obtained by our fusion method, contain less noise and preserve more structure information and edge information from source images. The advantage of our algorithm is more obvious when the $N_{abf}$ is used to assess the fused images. 

Although the $FMI_{pixel}$ for our fused images are not the best, its values are still very close to the best one, and the results obtained by our method to improve the fusion performance in term of $N_{abf}$, $SSIM_a$ and $EPI_a$ are acceptable.

\section{Conclusions}
\label{con}
In this article we have proposed a novel fusion algorithm based on ResNet50 and ZCA operation for infrared and visible image fusion. Firstly, the source images are directly fed into ResNet50 network to obtain the deep features. Following this ZCA operation which is also called whitening, is used to project the original deep features into a sparse subspace. The local average $l_1$-norm is utilized to obtain the initial weight maps. Then bicubic interpolation is used to resize initial weight maps to the source images size. A soft-max operation is used to obtain the final weight maps. Finally, the fused image is reconstructed by weighted-average strategy which combines the final weight maps and source images. Experimental results show that the proposed fusion method has better fusion performance in both objective and subjective evaluation.

\end{document}